\documentclass{llncs}
\usepackage[latin1]{inputenc}
\usepackage{epsfig}
\usepackage{tabularx}
\usepackage{url}
\begin{document}

\pagestyle{headings}

\mainmatter

\title{
MultiKulti Algorithm: Migrating the Most Different Genotypes in an Island Model
}

\titlerunning{Lecture Notes in Computer Science}
\author{Lourdes Araujo$^1$, Juan J. Merelo Guerv\'os$^2$, Carlos Cotta$^3$ and Francisco Fern\'andez de Vega$^4$}
\authorrunning{L.\ Araujo \& J.J.\ Merelo \& C. Cotta \& F.Fern\'andez}
\institute{$^1$Dpto. Lenguajes y Sistemas Inform\'aticos. UNED, Madrid 28040, Spain, \email{lurdes@lsi.uned.es}\\
$^2$Departamento de Arquitectura y Tecnolog\'ia de Computadores, 
Universidad de Granada, Granada 18071, Spain, \email{jmerelo@geneura.ugr.es},\\
$^3$Departamento de Lenguajes y Ciencias de la Computaci\'on,
Universidad de M\'alaga, Spain, \email{ccottap@lcc.uma.es},\\
$^4$Departamento de Arquitectura y Tecnolog\'ia de Computadores, 
Universidad de Extremadura, Spain, \email{fcofdez@unex.es}
}

\maketitle
\begin{abstract}
Migration policies in distributed evolutionary algorithms has not been
 an active research area until recently. However, in the same way as
 operators have an impact on performance, the choice of migrants is
 due to have an impact too. In this paper we propose a new policy (named {\em multikulti}) for
 choosing the individuals that are going to be sent to other nodes,
 based on {\em multiculturality}: the individual sent should be as
 different as possible to the receiving population (represented in several possible ways). We have checked this
 policy on different discrete optimization problems, and found that, in
 average or in median, this policy outperforms {\em classical} ones like
 sending the best or a random individual. 
\end{abstract}


\section{Introduction and state of the art}

Evolutionary algorithms (EAs) make individuals in the population
evolve in parallel, which suggests that the exploitation of
parallelism can be quite natural in these algorithms.  This has led to
many efforts to parallelize them with the intention of reducing the
execution time and improving the quality of the solutions.  There have
been several attempts to classify the numerous works devoted to
parallel EAs \cite{CantuBook00,Konfrst04}, with {\em island} or {\em
  coarse grained} models being one of the most popular approach of
parallelization of EAs.  This approach is usually implemented on
distributed systems, since it does not require a high cost in
communications:
nodes only exchange a few individuals after several generations.  The
population is divided in subpopulations, which usually evolve
isolatedly except for {\em migrations}, that is, the exchange of some
individuals after a number of generations.  The behavior of the island
model differs from the sequential EA, since the composition, and thus
the dynamics, of every subpopulation is different: since
subpopulations are smaller than the whole population, the parallel EA
will converge faster. Furthermore, migrations among subpopulations
usually improve the quality of the sequential solutions
\cite{AlbaNT02}, which makes the parallel model interesting even for
sequential executions.
Not only does this apply to genetic algorithms; similar results have been found for Genetic Programming: Tomassini et
al. \cite{tomassini:2003:EA} have analyzed diversity in
multipopulation genetic programming (GP) finding a correlation between
diversity and the better convergence properties of distributed GP.

These results have inspired this work. 
Diversity in the subpopulation is so important that it leads to
improvement in quality and efficiency at the same time.
Accordingly, we have looked for the way of enhancing the diversity 
induced by the usual migration policies, by using the notion of {\em multiculturality}: migrants should be chosen on the basis of genotypic difference to the receiving population; we have called this new migrant-selection policy {\em multikulti}\footnote{Multikulti, as defined by the wikipedia entry (\url{http://en.wikipedia.org/wiki/Multikulti}), is an slogan for a multicultural approach to public policy}.

Let us then look at how these diversity enhancement could be realized
through migration policies, which include several issues:
\begin{itemize}
\item
the number of individuals undergoing migration,
\item
the frequency of migration, i.e. the number of generations or evaluations
between migrations,
\item
the policy for selecting migrants,
\item
the migration replacement policy,
\item
the topology of the communication among subpopulations,
\item
the synchronous or asynchronous nature of the communications.
\end{itemize}
These issues have been investigated in different papers:
Alba et al. \cite{AlbaT01}
compare synchronous and asynchronous migration policies, 
Herrera et al. \cite{herrera99} studied some of the aforementioned issues in a hierarchical 
configuration of subpopulations,
and Cant\'u-Paz \cite{cantu99migration,cantu01migration}, Alba and Troya \cite{AlbaT00}, 
and Noda et al. \cite{noda02} have analyzed different migration policies. 
Several results presented in these mentioned works indicate that diversity is
a fundamental key in the success of the island model. For example,
works comparing synchronous versus asynchronous models \cite{AlbaT01},
have found that the asynchronous algorithms outperformed the synchronous ones
in all the experiments. 
Cant\'u-Paz \cite{cantu99migration,cantu01migration} has studied the four 
possible combinations of random and fitness-based emigration and replacement
of existing individuals. He found that 
the migration policy that causes the greatest reduction in work (takeover time\footnote{Number of generations required to converge to the best individual from the initial
population, by applying selection only}) 
is to choose both the
migrants and the replacements according to their fitness, because
this policy increases the selection pressure and may cause the
algorithm to converge significantly faster.
However, if convergence is too fast it can lead to algorithm failure, 
as Cant\'u-Paz \cite{cantu01migration} states referring to parallel EAs: 
\begin{quotation}
{\em Rapid convergence is desirable, but an excessively fast convergence may
cause the EA to converge prematurely to a suboptimal solution.}
\end{quotation}
So, other policies must also be considered.
In fact, Alba and Troya \cite{AlbaT00} found that 
migration of a random string prevents the ``conquest'' effect in the target
island for small or medium sized sub-populations.
Noda et al. \cite{noda02} have proposed choosing which
 individuals to migrate and/or replace adaptively depending on some knowledge-oriented
rules. To do this, each agent receives information about the fitness
function from its peers. The tested adaptive policies have proved useful
providing best solutions than the sequential execution.

In spite of the results shown above, there is still a number of issues that have not been investigated yet.
In this work we also focus on the policy for migrants selection.
Previous works dealing with this aspect have studied the use of any
of the selection operators usually applied in evolutionary algorithms: proportional
selection, tournament, random, etc.
Only the work by Noda et al. \cite{noda02} considers, among other policies, one in which
the individuals sent are chosen to be quite different from others previously sent.

\begin{figure}[tbh]
\centerline{\includegraphics[width=8cm,clip=]{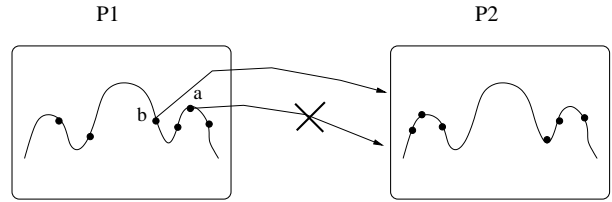}}
\caption{Illustration of the {\em multikulti} migrant selection policy: Individual
  {\sf a} in subpopulation P1 has a higher
fitness than individual {\sf b}. However {\sf b} is more different to
the best individual in subpopulation P2.
Accordingly, {\sf b} is selected to be sent in the migration. 
} 
\label{fidea}
\end{figure}

The aim of this work is to exploit differences in the various subpopulation.
To do this, we focus on the selection of the individuals to be sent to other subpopulation.
Our thesis is that migrating individuals different
enough to the destination subpopulation instead of the best
individuals can result in a better performance through the diversity
enhancement it produces.
Consider the Figure \ref{fidea} with two subpopulations. The black points represent
the distribution of the population along the function to optimize.
Individual {\sf a} in subpopulation P1 has the highest fitness, and thus it would be
sent to subpopulation P2 following the most common migration policies.
We propose to send individual {\sf b}, whose genotype is quite different from those of
subpopulation P2. In the example it would lead to exploration of a new area of the search
space where the global optimum is placed. 
In order to achieve this, the process corresponding to subpopulation P1 needs to receive 
information on the composition of the individuals in subpopulation P2.
We have considered different ways of providing this information in a concise manner. 
One of them is taken the best individual of subpopulation P2 as representative.
The other one is using a kind of average genotype, the consensus sequence
described later, as representative of subpopulation P2.
Another important issue to investigate if the trade-off between promoting diversity
and favoring the best individuals. The risk of sending the most different individual
as migrant is that if its fitness value is low compared to those of the destination
subpopulation, the migrant would probably disappear immediately.
Therefore, another question tested in the experiments has been if the most different
individual is fit enough to survive when migrating, or if it is best to select the
most different from an elite.

The rest of the paper proceeds as follows:
section 2 describes the model details;
section 3 is devoted to describe the evolutionary algorithm and its implementation;
section 4 presents and discusses the experimental results,
and section 5 draws the main conclusions of this work.

\section{Model Description}

\begin{figure}[bth]
\centering
\centerline{\includegraphics[width=8cm,clip=]{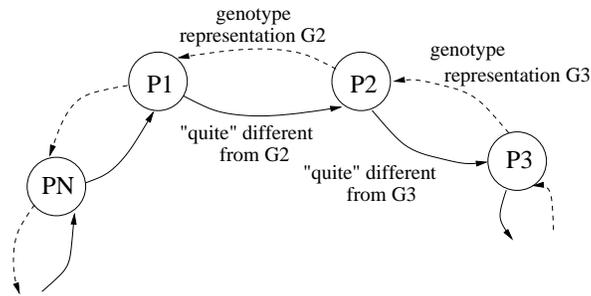}}
\caption{Scheme of the multikulti algorithms.}
\label{fmodel}
\end{figure}

We have considered a ring topology (Figure \ref{fmodel}), in which
each node can send one or more individuals to the next node in the ring. 
To perform the choice of the migrants, the node $P_i$ receives from node $P_{i+1}$ information
about the genotype of its subpopulation. 
We have considered two different ways of representing this information in a concise manner:
\begin{enumerate}
\item
With the best individual of the subpopulation. After a number of generations without exchanging individuals
we can expect that each subpopulation is close enough to converge for the best individual being
a fair representation of the whole population.
\item
With the consensus sequence of the population.
This is a concept taken from biology where it is defined as the sequence that reflects the most common choice of 
base or amino acid at each position of a genome. Areas of particularly good agreement often 
represent conserved functional domains.
In our case it is composed of the most frequent alelo for each
position of the genotype.
\end{enumerate}
Once the node $P_i$ has got this information, it sends to node $P_{i+1}$ an individual
different enough from the subpopulation $P_{i+1}$ representative. Here we have considered two approaches: 
\begin{itemize}
\item
{\em Base}: Selecting the most different from the subpopulation $P_i$.
\item
{\em Elite}: Selecting the most different among  the best half of subpopulation $P_i$.
\end{itemize}

\section{Implementation}

Chromosomes \cite{Holland} of our GA are fixed-length binary strings.
The selection mechanism to choose individuals for the new population
uses a steady state algorithm, with two-point crossover operator and
single-bit-flip mutation. The
rest of the parameters are shown in tables \ref{tab:parameters}(a) 
and \ref{tab:parameters}(b). 

\begin{table}[htbp]
\centering
\begin{tabular}{c|c}
\begin{minipage}[t]{5.5cm}
\begin{tabular}{l|c}
Parameter & Value \\
\hline
Population & 32 \\
Selection rate & 60\%\\
Generations to migration & 20 \\
Mutation priority & 2 \\
2-point crossover priority & 3 \\
\end{tabular}
\end{minipage}
&
\begin{minipage}[t]{5.5cm}
\begin{tabular}{l|c}

          Parameter & Value \\
\hline
Chromosome length & 120\\
Population & 256 \\
Selection rate & 20\%\\
Generations to migration & 20 \\
Mutation priority & 2 \\
2-point crossover priority & 3 \\
Max number of evaluations & 200000 \\
\end{tabular}
\end{minipage}\\
(a) & (b)\\
\end{tabular}
\caption{Evolutionary algorithm parameters used in the P-Peaks (a) and in the
\textsf{MMDP}(b) experiments.
\label{tab:parameters}
}
\end{table}

{\sf P-Peaks} and the massively multimodal
deceptive problem ({\sf MMDP}), two of the three 
discrete optimization problems presented by Giacobini et al. in
\cite{giacobini:evocop06} have been selected for testing. These
problems, while being both multimodal, represent different degrees of
difficulty for parallel evolutionary optimization. They will be
described below.

These two problems have been implemented and integrated in the
\texttt{Al\-go\-rithm::Evo\-lu\-tio\-na\-ry} library, which is freely available
under the GPL license from
\url{http://opeal.cvs.sourceforge.net/opeal/Algorithm-Evolutionary/}. In order to simulate a 
parallel algorithm, the {\em cooperative multitasking} Perl module
\textsc{POE} has been used; each node is represented by a POE {\em
  session}. The rest of the
evolutionary algorithm has been implemented using the same
\texttt{Al\-go\-rithm::Evo\-lu\-tio\-nary} Perl module \cite{ecperl}. The program,
along with the parameter sets used, is also available under an open source
license from the same site.

In this simulated parallel scenario, each node runs a rank-based
substitution, steady state algorithm. At the end of a preset number of
generations, each node sends a single individual to the other node
according to the policy being tested

\subsection{Problems tested}

Two functions have been used for testing: {\sf P-Peaks} and \textsf{MMDP}, two of the three discrete optimization problems presented in \cite{giacobini:evocop06}: The massively multimodal
deceptive problem ({\sf MMDP}) and the problem generator {\sf P-Peaks}. These
problems, while being both multimodal, represent different degrees of
difficulty for parallel evolutionary optimization. They will be
described next.

The {\sf MMDP} \cite{goldberg92massive} is a deceptive problem compose of $k$ subproblems of 6 bits each one ($s_i$). Depending of the number of ones (unitation) $s_i$ takes the values depicted next:\\
$fitness_{s_i}(0) = 1.0$, $fitness_{s_i}(1) = 0.0$, $fitness_{s_i}(2) = 0.360384$,\\ $fitness_{s_i}(3) = 0.640576$
$fitness_{s_i}(4) = 0.360384$, $fitness_{s_i}(5) = 0.0$,\\ $fitness_{s_i}(6) = 1.0$

The fitness value is defined as the summatory of the $s_i$ subproblems with an optimum of $k$ (equation \ref{eq:mmdp}).
The number of local optima is quite large ($22^k$), while there are only
$2^k$ global solutions. In this paper, we consider a single instance with $k={20}$.
\begin{equation}\label{eq:mmdp}
f_{MMDP}(\vec s)= \sum_{i=1}^{k} fitness_{s_i}
\end{equation}
The {\sf P-Peaks} problems is a multimodal problem generator
proposed by De Jong in  \cite{dejong97using}, and is created by generating $P$ random $N-bit$ strings where the fitness value of a string $\vec x$ is the number of bits that $\vec x$ has in common
with the nearest peak divided by $N$. In the experiments made in this
paper we will consider $P=100$; the optimum fitness is 1.0.
\begin{equation}\label{eq:ppeaks}
f_{P-Peaks}(\vec x)= \frac{1}{N} \max_{1 \leq i \leq p}\{ N - HammingDistance(\vec x,Peak_i)\}
\end{equation}
We consider an instance of $P=100$ and 64 bits where the optimum fitness is 1.0 (Equation \ref{eq:ppeaks}).

These two problems have been also implemented and integrated in the
\texttt{Al\-go\-rithm::Evo\-lutio\-nary} library.

\section{Experimental results}

\begin{figure}[tbh]
\begin{center}
\includegraphics[scale=0.5]{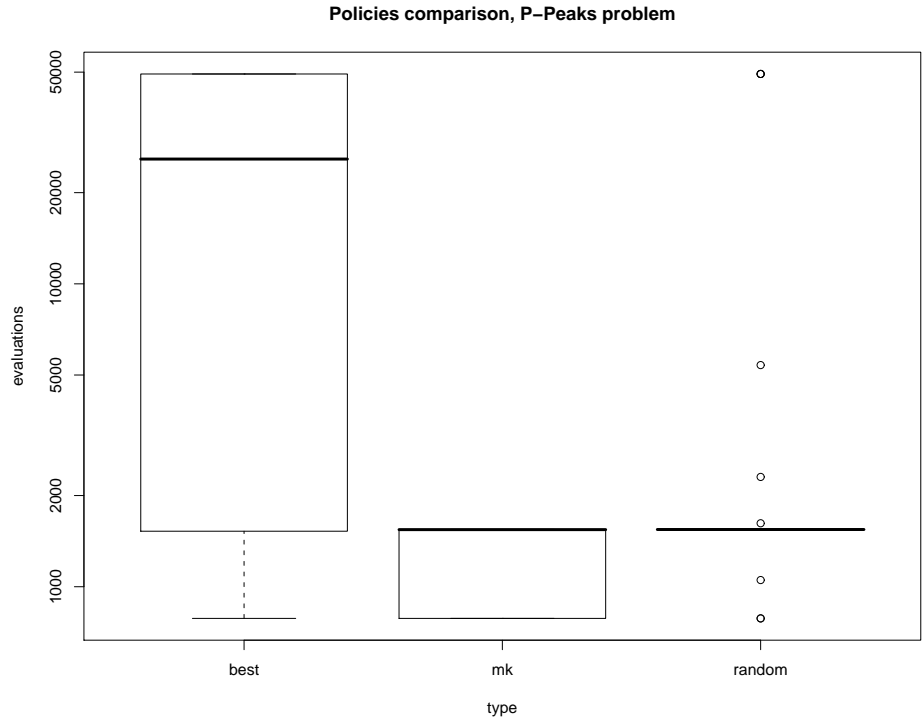}
\caption{Boxplot (with logarithmic $y$ axis) of the number of evaluations needed to find the
 solution in the {\sf P-Peaks} problem. {\em best} represents the behavior of
 the experiment when the best individual in the population is sent, {\em
 random} with a random individual, and finally {\em mk}, in the middle,
 stands for {\em multikulti}, the algorithm we are testing in this paper
 which sends the individual in the population most different to the best
 in the receiving population.  
\label{fig:evals:p-peaks}}
\end{center}
\end{figure}

First, we tested several parameter configurations for the {\sf P-Peaks}
problem. In general, when diversity conditions are not too harsh, the
performance difference between different migration policies is not too
high. Eventually, when the going gets tough, differences such as those
shown in Figure \ref{fig:evals:p-peaks} do appear. In that graph, taken
for a 8-node, population = 32 experiment, results are quite different
depending on the migration policy. For starters, sending the best
individual yields the worst results. If we consider the median, the {\em multikulti}
 is similar to the {\em random} policy, but its behavior is better if we
 consider the average and the worst case, as shown in table \ref{tab:stats}.
\begin{table} [htbp]
\centering
\begin{tabular}{l|c|c}
Policy & Median & Average \\
\hline
Best & 25820 & 25310\\
Random & 1545 & 6410 \\
Multikulti & 1544 & 1252 \\
\end{tabular}
\vspace{10pt}
\caption{Statistics for number of evaluations of different migration
  policies for the {\sf P-Peaks} problem.\label{tab:stats}}
\end{table}

\begin{figure}[tbh]
\begin{center}
\includegraphics[scale=0.6]{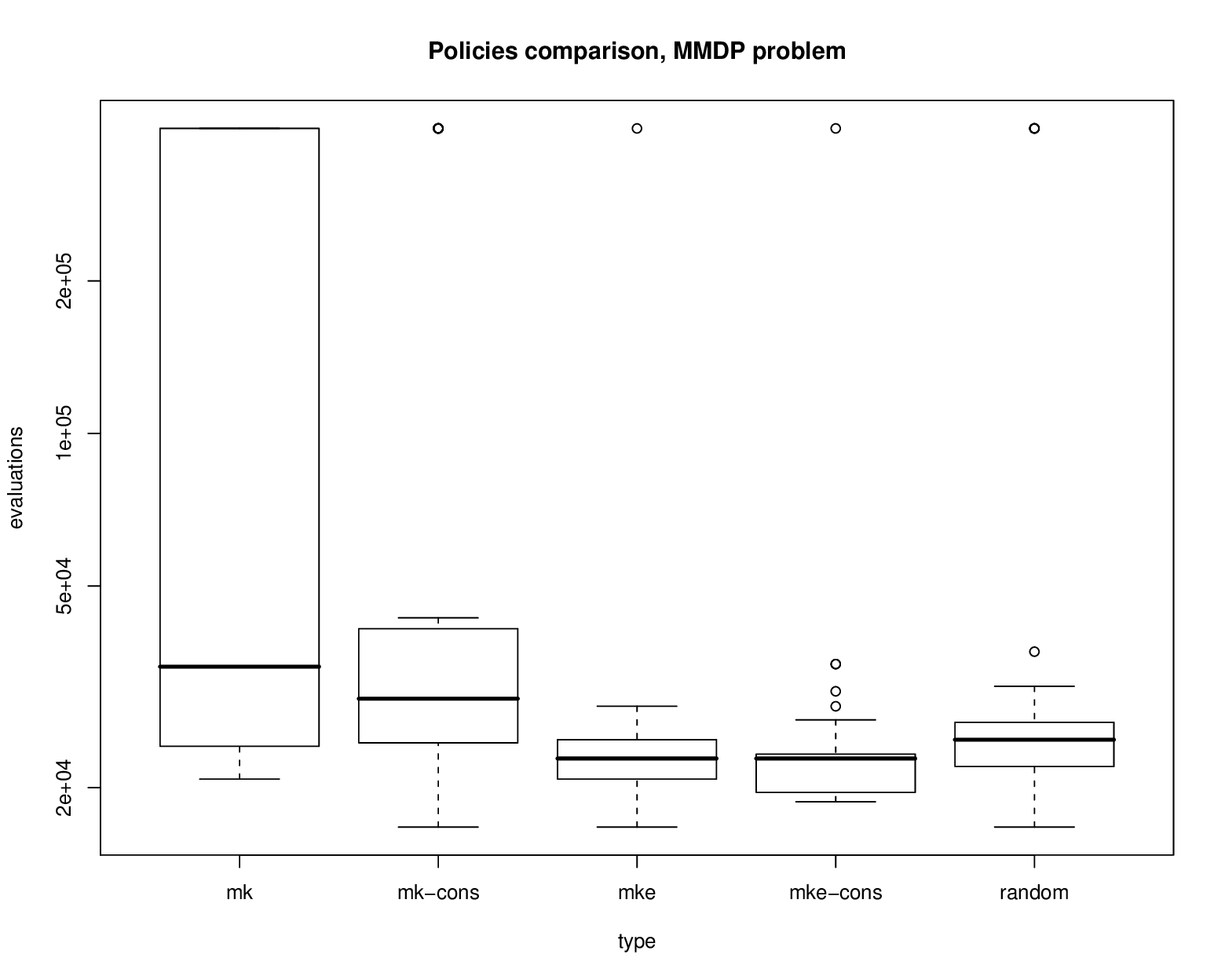}
\caption{Boxplot (with logarithmic $y$ axis) of the number of evaluations needed to find the
 solution for {\sf MMDP}. Three different versions of the multikulti migration
 policy have been tested here: {\sf mk}, in which the most different
 individual is sent; {\sf mk-cons}, that sends the most different to the target's consensus string, 
{\sf multikulti-elite} ({\sf mke}) which
 chooses the most different among the 50\% most fit,
 and {\sf mke-cons} which sends the  most different to the target's consensus string among the 50\% most fit.
The rightmost result corresponds to random migration.\label{fig:evals:mmdp}}
\end{center}
\end{figure}

The {\sf MMDP} was also tested with a similar setup; results are shown
in figure \ref{fig:evals:mmdp}
where three new versions of the algorithm, named {\sf multikulti-elite}, {\sf consensus-multikulti}
and {\sf multikulti-elite-consensus},
have been tested. {\sf Multikulti-elite}  chooses the individual most different to
the receiving population best individual, but only among the 50\%
best. In this case, not surprisingly, this strategy beats the
multikulti by far as well as the random strategy, but more
closely. This is probably due to the nature of the {\sf MMDP}: it is a
deceptive problem, where increasing diversity might not have the
desired result, since competing conventions for the same 6-bits
portion can lead in each population. This why, instead of sending the
most different as the multikulti policy does, doing it with one that
is different enough, but at the same time, fit enough, produces the
best results.

The second strategy tested, {\sf consensus-multikulti}, achieves an
intermediate performance among the two, beating the multikulti, but
achieving worse results than {\sf multikulti-elite}. This strategy
sends the individual that is most different to the consensus string,
that is, the string whose every bit represents the majority value for
that position among the population. The result is probably due to the
same reason: a value that is too different represents a high
disturbance, and thus is bad for diversity. In fact, the third strategy,
{\sf multikulti-elite-consensus}, which sends the individual 
most different to the consensus string which is  among the 50\%
best, achieves results similar to {\sf Multikulti-elite}, although, in
general, a bit better.
 
\begin{figure}
\begin{center}
    \begin{tabular}{ccc}
      \includegraphics[scale=0.16]{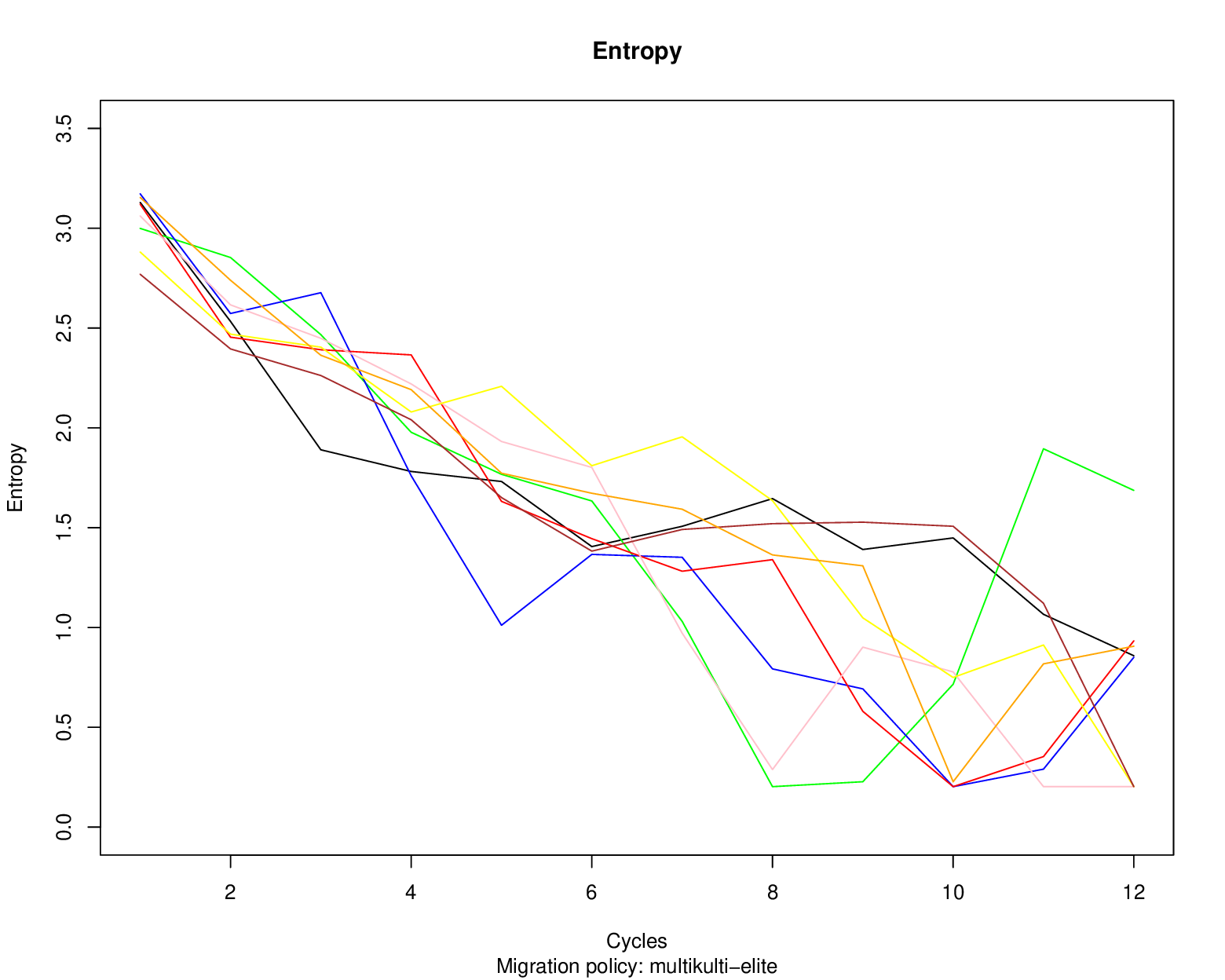} &
      \includegraphics[scale=0.16]{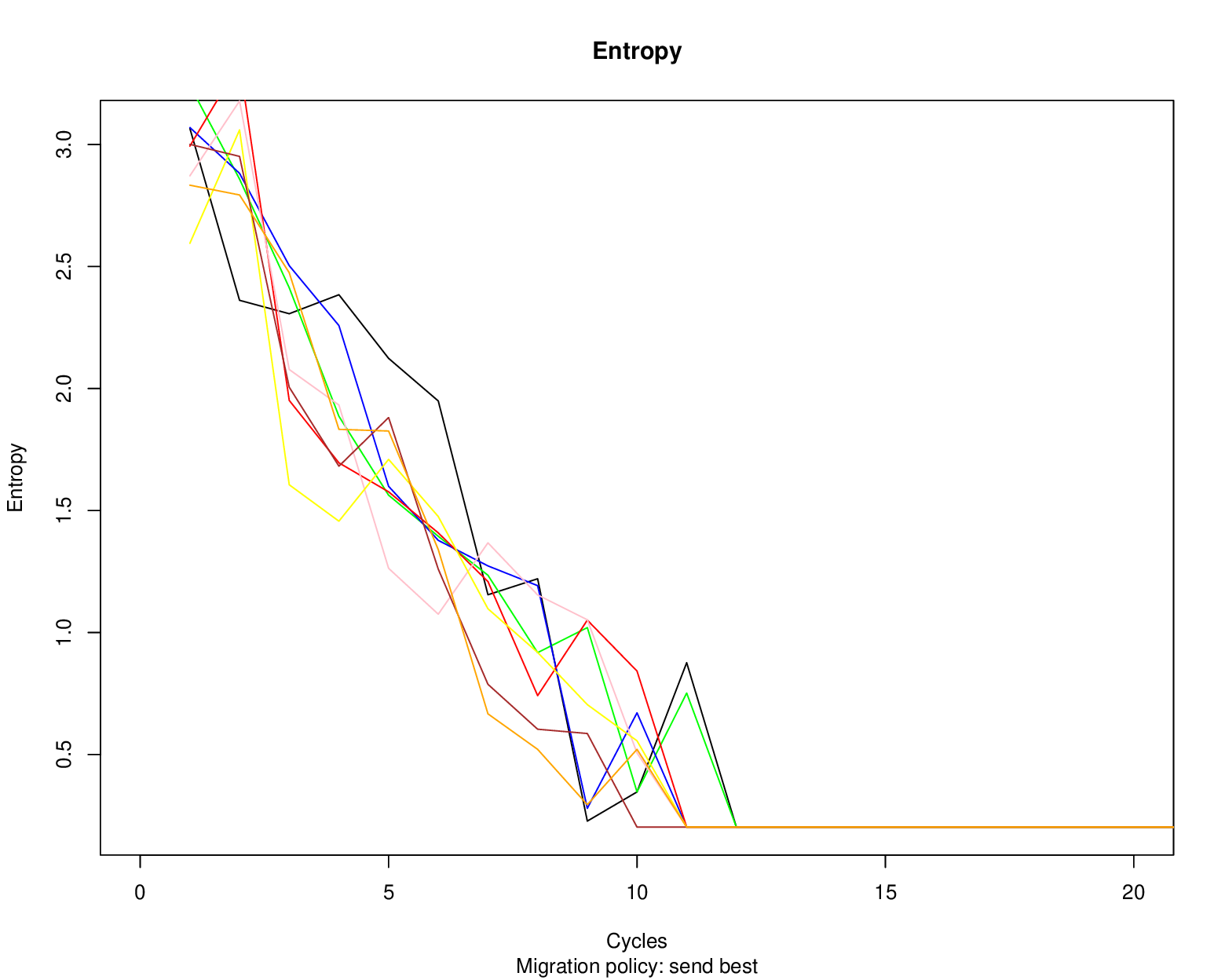}& 
      \includegraphics[scale=0.16]{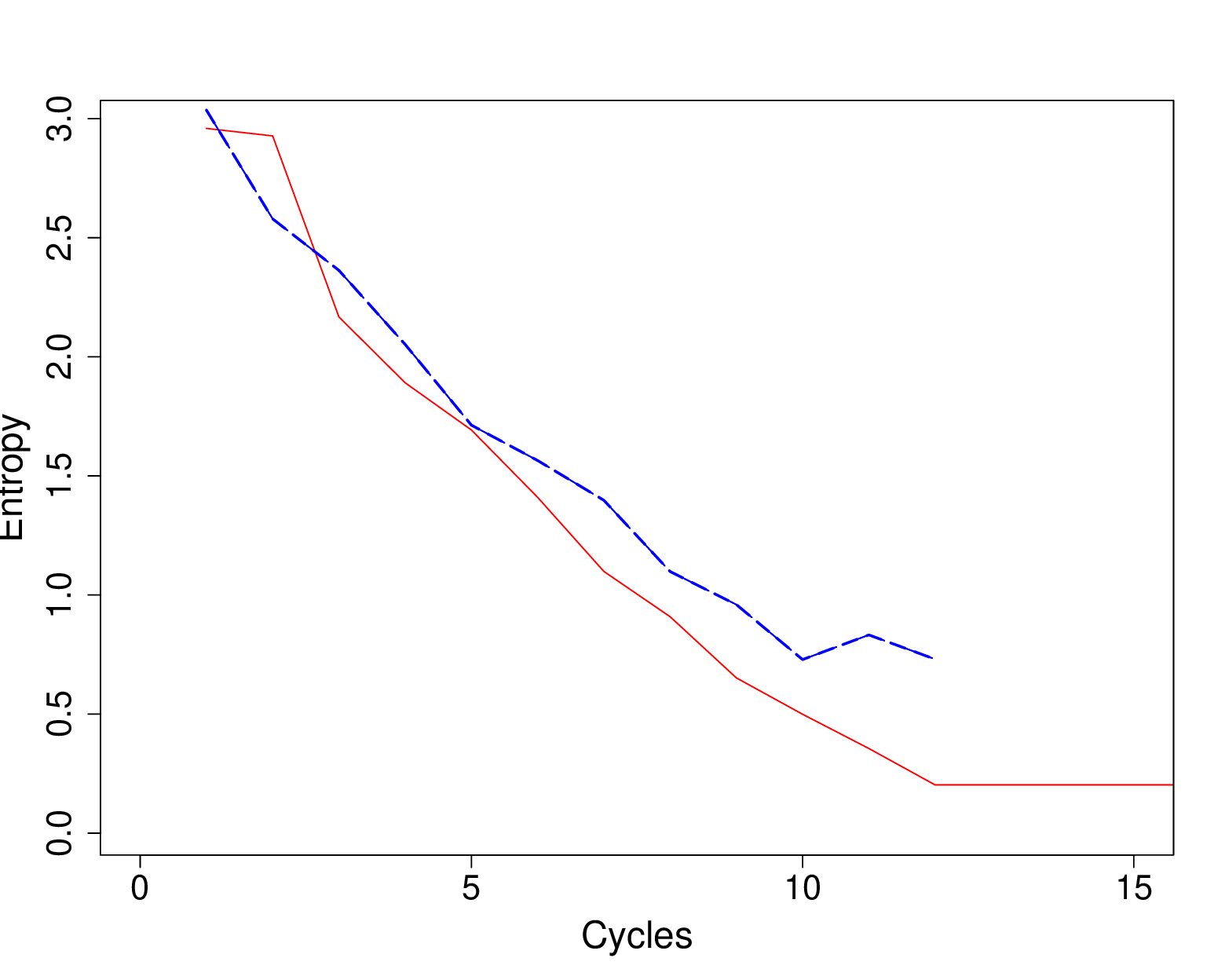} \\
      mke & best& average\\
    \end{tabular}
    \caption{Entropy (computed using the Shannon formula $H(P) = - \sum_{g \in P} p( f( g )) log_{b} p( f( g ) )$, 
     where $g$ is a member of the population, $f(g)$ its fitness, and $p(f(g))$ the frequency of 
     that fitness across the whole population) in a typical run of the
      \textsf{MMDP} problem, with
      the multikulti-elite migration policy (left) and the best migration policy (center).
      Every line corresponds to a different population, of the eight
      running in parallel.
      The figure on the right compares average values, with the
      dashed line corresponding to the multikulti-elite experiment and
      the other to the experiment that sends the best individual.
\label{fig:entropy}
}
\end{center}
\end{figure}

In order to investigate what is going on, we have measured the entropy for the {\sf MMDP}.
The results are shown in Figure \ref{fig:entropy},
which shows the different evolution paths of phenotypic entropy
(computed using the Shannon formula) with the
multikulti-elite migration policy (left) and the best migration policy (center).
The behavior es quite different. The multikulti policy, not only keeps
the entropy high, but considerably increases it in some populations
during evolution. The policy of migrating the best provides quite much
lower levels of entropy; with a decreasing trend that never changes,
leading to a collapse of entropy from cycle 12. This proves the
utility of the multikulti policy to maintain diversity, and supports
the result that the improvement in the number of evaluations is due
precisely to this diversity-enhancing effect brough by the {\em
  multikulti} policies.  

%

\section{Conclusions}

This paper has explored new alternatives to promote diversity in an island
model.  This is achieved by selecting as migrant individuals with a genotype 
different enough to the destination population.
Because there is a trade-off between promoting
diversity and favoring the best individuals, we have performed experiments
to find out the degree of difference which produces the best results.
These experiments have shown that results only improve substantially
if the migrant is chosen from the elite; that is, those with an above-average fitness, which indicates that diversity
only improves the results if the incoming migrant has a minimum level of quality.
We have compared two different ways of characterizing the destination population:
the best individual and the consensus sequence. Results have shown that
both of them represent appropriately the population, with the consensus
sequence performing only slightly better. Studying the phenotypic entropy of the
population we have found that our method effectively improves
entropy by avoiding entropy to fall too fast and also creating an entropy differential among populations, and thus diversity.

In the future we intend to develop a parallel implementation of the system,
which will allow us to measure also execution times.
We are also working on alternative mechanisms to characterize the
destination population, and thus select the more appropriate
migrants. We will also test results obtained by changing other
algorithm parameter such as number of migrants or the number of nodes.  

\section*{Acknowledgements}

This paper has been funded in part by the Spanish MICYT project NoHNES
(Spanish Ministerio de Educaci\'on y Ciencia - TIN2007-68083) and the
Junta de Andaluc\'ia P06-TIC-02025.

\bibliographystyle{splncs}
\bibliography{parallelAG,GA-general,asynchronous,geneura}

\end{document}